\documentclass[runningheads]{llncs}
\usepackage[T1]{fontenc}
\usepackage{multirow}
\usepackage{graphicx}
\usepackage{amsmath,amssymb}
\usepackage{cleveref}
\begin{document}
\title{ALS-HAR: Harnessing Wearable Ambient Light Sensors to Enhance IMU-based Human Activity Recogntion}
\titlerunning{ALS-HAR}
\author{Lala Shakti Swarup Ray\inst{1}\orcidID{0000-0002-7133-0205}\thanks{lala\_shakti\_swarup.ray@dfki.de} \and
Daniel Geißler\inst{1}\and
Mengxi Liu\inst{1}\and
Bo Zhou\inst{1,2}\and
Sungho Suh\inst{1,2}\and
Paul Lukowicz\inst{1,2}}
\authorrunning{Ray et al.}
\institute{German Research Center for Artificial Intelligence, Kaiserslautern, Germany \\
\and
RPTU Kaiserslautern-Landau, Kaiserslautern, Germany\\}
\maketitle              % typeset the header of the contribution
\begin{abstract}
Despite the widespread integration of ambient light sensors (ALS) in smart devices commonly used for screen brightness adaptation, their application in human activity recognition (HAR), primarily through body-worn ALS, is largely unexplored. 
In this work, we developed ALS-HAR, a robust wearable light-based motion activity classifier. 
Although ALS-HAR achieves comparable accuracy to other modalities, its natural sensitivity to external disturbances, such as changes in ambient light, weather conditions, or indoor lighting, makes it challenging for daily use. 
To address such drawbacks, we introduce strategies to enhance environment-invariant IMU-based activity classifications through augmented multi-modal and contrastive classifications by transferring the knowledge extracted from the ALS.
Our experiments on a real-world activity dataset for three different scenarios demonstrate that while ALS-HAR's accuracy strongly relies on external lighting conditions, cross-modal information can still improve other HAR systems, such as IMU-based classifiers.
Even in scenarios where ALS performs insufficiently, the additional knowledge enables improved accuracy and macro F1 score by up to 4.2 \% and 6.4 \%, respectively, for IMU-based classifiers and even surpasses multi-modal sensor fusion models in two of our three experiment scenarios. 
Our research highlights the untapped potential of ALS integration in advancing sensor-based HAR technology, paving the way for practical and efficient wearable ALS-based activity recognition systems with potential applications in healthcare, sports monitoring, and smart indoor environments. 
\keywords{Human Activity Recognition \and Ambient Light Sensor \and Sensor Fusion \and Contrastive Learning \and IMU Sensing}
\end{abstract}
\section{Introduction}
Sensor-based HAR has gained increasing interest in research and industry over the past decade, advocating various sensor modalities like pressure sensors \cite{zhou2022quali,ray2023pressim}, EMG sensors \cite{mekruksavanich2023human,martin2020electromyographic}, impedance sensors \cite{liu2024imove,jiang2020hand} and capacitive sensors \cite{kumar2023tracking}.
Especially with the ubiquity and availability of smart devices, embedded sensors like inertial measurement units (IMUs) \cite{li2019adaptive,mcgrath2021upper,pesenti2023imu} have gained popularity due to their ability to capture motion-related data accurately through great information density.

However, despite the extensive exploration of IMUs in HAR, there is a growing trend towards investigating the potential of other embedded sensors like BLE \cite{demrozi2021estimating,vesa2020human}, WiFi signals \cite{zhang2020data,yadav2022csitime}, temperature sensors \cite{demrozi2020human} and ALS \cite{xu2022visual}, which is presented in this work.
Nowadays, ALS is embedded in almost all portable smart devices with a screen primarily used for adaptive screen brightness adjustments based on changing environmental lighting conditions \cite{isuwa2023maximising}.
Such a sensor operates passively without direct user interaction and can be exploited to provide valuable contextual information about the user's surroundings and activities.
Additionally, it consumes minimal power, contributing to energy-efficient implementations, particularly on battery-powered devices.
For this work, we aim to benefit from the ubiquity of ALS in smart mobile devices, eliminating the need for additional hardware through straightforward accessibility. 

To the best of our knowledge, despite the promising advantages and availability, limited research has been done regarding exploring body-worn ALS in HAR as a motion-sensing modality. 
Throughout related Multi-modal sensor fusion works like \cite{light202,shi2021intelligent,s24113367}, static ALS and other ambient sensors placed in the environment are used for positional understanding for motion localization. 

Wearable, body-worn ALS holds promise for HAR, particularly in indoor environments with stable lighting conditions where external factors affecting light intensity are minimal. 
Xu et al. have exploited the wearable ALS to generate IMU data for improving nursing-based HAR \cite{xu2022visual}. Similarly, Sadaghiani et al. used wearable photodiodes to gather blood pressure signals of the wearer's body \cite{sadaghiani2023ambient}.
Despite the promises, these models suffer from the obvious problem of being sensitive to external lighting conditions. When the changes in light conditions are more significant or comparable to those impacted by the user's movement, the model performance drastically decreases, making them useless for such conditions. 
Even further, just like the overall nature of vision sensors, the ALS can not work in dark environments \cite{xu2022visual} as stated by Xu et al.

\begin{figure}[t]
\centering
\includegraphics[width=1\linewidth]{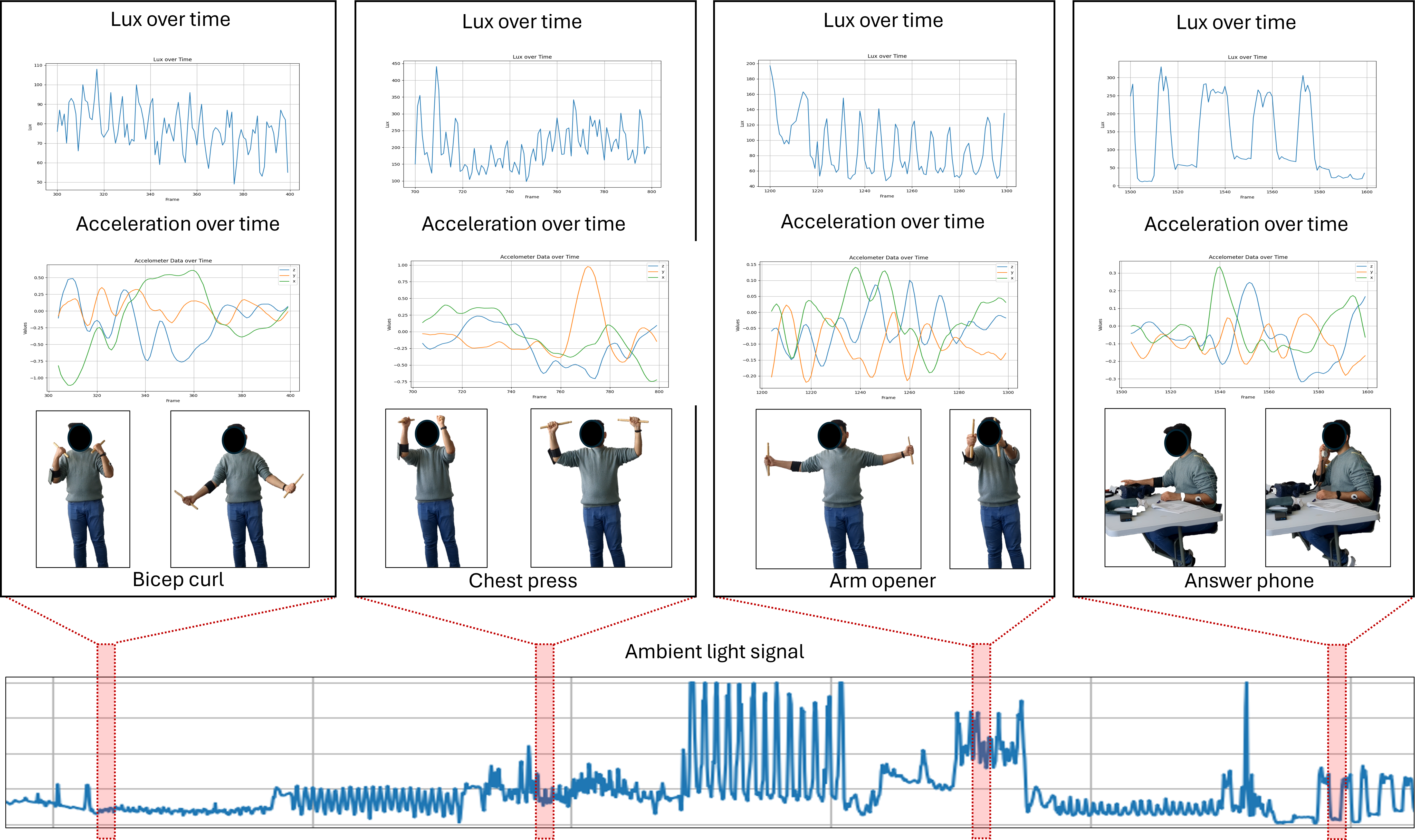}
\caption{Synchronized ambient light and IMU accelerometer signals extracted from the collected dataset aligned with the some of the labelled activity classes.}
\label{fig:dataset}
\end{figure}

In this paper, we try to solve this problem by investigating different cross-modal approaches that can empower other sensor-based modalities, like IMU-based HAR, by using knowledge transfer techniques from ALS to IMU.
Therefore, we aim to maximize the knowledge extracted from ALS even in unfavorable, fluctuating light conditions, improving HAR performance through ALS independently of environmental influences.
In summary, the main contribution of this work can be summarized as follows:
\begin{itemize}
    \item Multi-modal HAR dataset: A novel multi-modal dataset containing nine different activities for a total of 5.28 hours performed by 16 participants along three different environmental scenarios gathering right wrist IMU signal, right wrist ALS signal, video footage and SMPL pose synchronized together as visualized in Figure \ref{fig:dataset}. 
    \item LightHAR: An activity recognition model based only on the wrist-based ALS with a detailed comparison to wrist-based IMU, 3D pose-based, and video-based activity recognition for the three different scenarios.
    \item Light embedded InertialHAR: Two different strategies to improve inertial HAR utilizing both ALS and IMU signal during training and only IMU signal for inference.
\end{itemize}

\section{Related work}
\subsection{ALS-HAR}
HAR is a continuously evolving field that leverages various sensing modalities to identify and monitor human activities. 
ALS has shown promise in enhancing the accuracy and applicability of HAR systems, especially deployed as external environmental sensors through works like \cite{light202} presenting a Deep Convolutional Neural Network to recognize human activities using binary ambient sensors to identify activities of daily living.
In \cite{ahamed2020internet}, the landscape of available sensors for HAR has been analyzed by Ahamed et al., investigating the importance of environmental sensors, especially the ALS, to detect the early signs of dementia in residential care.
Integrated into smart wallpapers, multi ALS has been implemented by Shi et al. to recognize human motions with an accuracy of 96\% utilizing the information of light reflections gathered through photodiodes hidden inside the wallpaper \cite{shi2021intelligent}.
Focusing on industrial scenarios and ambient assisted living, Salem et al. have proven the feasibility of fusing the sensor data from IMU and ALS to achieve an activity recognition performance of 90\% across a small set of three classes for each scenario \cite{Salem2022Improved}.

Environmental Sensing for HAR commonly possesses drawbacks on adaptation to changing light conditions and occlusion of the covered area, wherefore body-worn ALS can be an alternative to enhance HAR performance \cite{Lara2013A}.
Due to their simplicity in operation and low power consumption, they are commonly deployed in consumer wearable devices \cite{perez2021recent}.
In \cite{li2018self}, the benefits of low-power ALS have been deployed to harvest energy through photodiodes and simultaneously utilize the ambient light data for self-powered and robust finger gesture detection.
Similar work has been done through OptoSense, presenting a novel approach for developing body-worn ALS that is self-powered and capable of being integrated ubiquitously by leveraging photovoltaic cells both to power the sensors and to sense the ambient light, enabling the creation of energy-efficient HAR through light sensors \cite{zhang2020optosense}.
Similar to the approach of this work, Wang et al. present a multimodal feature fusion model utilizing geomagnetic, ALS, and accelerometer data collected from smartphones to enhance health activity monitoring accuracy in indoor environments by 13.65\% compared to classic sensor classification \cite{s24113367}.

However, the presented literature barely investigates changing environments and lighting conditions, commonly working in clean and optimal indoor environments, restating the motivation of this work.

\subsection{Knowledge Transfer in Sensor HAR}
Methods like Don't Freeze \cite{dontfreeze}, Virtual Fusion \cite{nguyen2023virtual} and Contrastive Left-Right HAR\cite{nshimyimana2023contrastive} tried using IMU sensors at different positions to improve the overall accuracy of body-worn IMU at specific positions using contrastive learning.
Approaches like Multi$^3$Net\cite{rey2024enhancing} have tried to improve sensor-based HAR accuracy using other widely available modalities like 3D poses and text embedding.
i-Move \cite{liu2024imove} improved IMU-based HAR using bio-impedance data through contrastive learning. 

We believe that the most important use case for ALS data would be empowering other sensor modalities that are more environment invariant through data collected in ideal environments, which we try to achieve through this work.

\section{Data Collection}

Our experiment encompassed three different scenarios based on different environmental and lighting conditions. It consisted of 16 participants doing 10 activities, including the Null class. The gender distribution was 5 females and 11 males, ages 24 to 35, and weights ranged between 53 kg and 88 kg. 

Scenario 1, consisting of subjects 1 to 10, was recorded in a controlled indoor environment with fixed lighting conditions. This environment is ideal for ALS-HAR because of the minimal interference of change in light due to external factors.
Scenario 2, consisting of subjects 11 to 13, was recorded in a relatively dark indoor environment with dynamic architectural lights. Most interference in lighting conditions is introduced due to these external factors rather than the motion of the subject itself, making it more challenging than the other two scenarios.
Scenario 3, consisting of subjects 14 to 16, was recorded in an outdoor environment during cloudy weather. The clouds and trees moved because wind created small interference in light signals, making it a practical dataset to showcase the usability of ALS-HAR.

Participants are engaged in a series of predetermined activities, including six distinct upper body fitness exercises \textit{boxing, biceps curls, chest press, shoulder, and chest press, arm hold and shoulder press,} and \textit{arm opener} sourced from Pamela's fitness routines available on YouTube (\footnote{https://www.youtube.com/@PamelaRf1}). Additionally, three supplementary hand-focused tasks \textit{sweeping a table, Answering the telephone,} and \textit{wearing a headset} were included, each lasting approximately 20 minutes.

We used existing consumer-grade devices for data collection to showcase the utility of ALS signals without facing the bottleneck of the new sensor introduction. 
We utilized a Samsung Galaxy S20 smartphone worn on top of the right wrist with a wristband facing outward, having the same relative position and orientation to the wrist irrespective of the user. This allowed for the collection of both light sensor and IMU data to fulfill the experimental requirements.
Data was collected using the Sensor Logger Android application (\footnote{https://github.com/tszheichoi/awesome-sensor-logger/}). 
Video recordings, captured using a back-facing camera of an iPhone SE, served as supplementary data for annotation purposes.

Sensor Logger automatically synchronizes light and IMU sensor, ensuring a consistent sampling rate of around 30 Hz throughout the session by taking a common time-stamp from the smartphone itself and matching the start and end of the session.
The videos collected by a separate smartphone are synchronized manually with the sensor data using a simple trick. 
At the start and end of each session, the subject needs to do the calibration movement, i.e., fold arms to touch both hands three times to make a unique pattern in the pose and the sensor signal. By mapping these unique patterns of the pose and the sensor signal, we can synchronize both together.
Afterward, the videos are manually annotated and can be used directly to annotate the sensor signals. 

Typical ALS available in smartphones uses a photodiode, a semiconductor device that generates an electrical current when exposed to light. The intensity of the current is proportional to the amount of light hitting the sensor.
The light signal, recorded in lux, is a unit of measurement for illuminance, representing the amount of light per unit area. In this context, lux provided insights into the ambient light conditions surrounding the experiment's environment, especially the light reaching the right wrist based on the subject's movement.

\begin{table}
\caption{Data statistics including subject mass (\(kilogram\)), height (\(centimeter\)), gender, and duration of the session (\(second\)) for the three different scenarios.}
\label{tab:stat}
\centering
\begin{tabular}{c|c|c|c|c|c|c}
\hline
Scenario &Subject ID & Age& Height(cm) & Weight(kg) & Gender & Duration(sec)\\
\hline
& 1 & 30 & 160 & 53 & Male & 1394\\
& 2 & 32 & 160 & 53 & Female & 1466\\
& 3 & 25 & 175 & 65 & Female & 1399\\
& 4 & 35 & 188 & 88 & Male & 1362\\
1: Indoor& 5 & 26 & 175 & 86 & Male & 1376\\
(Ideal)& 6 & 24 & 178 & 85 & Female & 1487\\
& 7 & 26 & 150 & 50 & Female & 1385\\
& 8 & 24 & 175 & 80 & Male & 1482\\
& 9 & 25 & 170 & 65 & Male & 1393\\
& 10 & 26 & 176 & 55 & Male & 1319\\
\hline
2: Indoor&11 & 27 & 187 & 85 & Male & 1225\\
(Challenging)& 12 & 30 & 160 & 53 & Male & 1240\\
&13 & 28 & 175 & 65 & Male & 1127\\
\hline
&14 & 26 & 176 & 55 & Male & 1185\\
3: Outdoor&15 & 35 & 168 & 65 & Male & 1305\\
&16 & 33 & 153 & 54 & Female & 1245\\
\hline
\end{tabular}
\end{table}

\section{Method}

\subsection{LightHAR}
\label{LightHAR}
We developed a robust ALS-based activity classifier tailored for light sensor data using a 1D bidirectional LSTM-based encoder architecture inspired by the DeepConvLSTM framework \cite{ordonez2016deep}. Our model processes input data $X$ with dimensions $(N, 1, 1)$, where $N$ represents the sequence length with unit feature dimension and a single channel. The architecture outputs a probability distribution over 10 classes (9 + null), denoted as $\hat{Y}$.

The model begins with a series of three 1D convolutional layers followed by batch-normalization and dropout layers, each designed for feature extraction. These layers sequentially process the input data to capture relevant patterns and characteristics from the light sensor signals. Each convolutional block is followed by a ReLU activation function, which introduces non-linearity into the model.
After feature extraction, the processed data is passed through a bidirectional LSTM layer to capture temporal dependencies in the sequence data. The bidirectional nature of the LSTM layers allows the model to consider both past and future information when making predictions, which enhances the overall performance of the activity classifier.
The output from the LSTM layers is then directed through dense layers for classification. The final layer outputs a probability distribution of the 10 classes, enabling the model to determine the most likely activity class for a given sequence of light sensor data.
Our architecture focuses on creating a lightweight model with robust and accurate classification based on light sensor inputs. 

\begin{figure}
\centering
\includegraphics[width=0.9\linewidth]{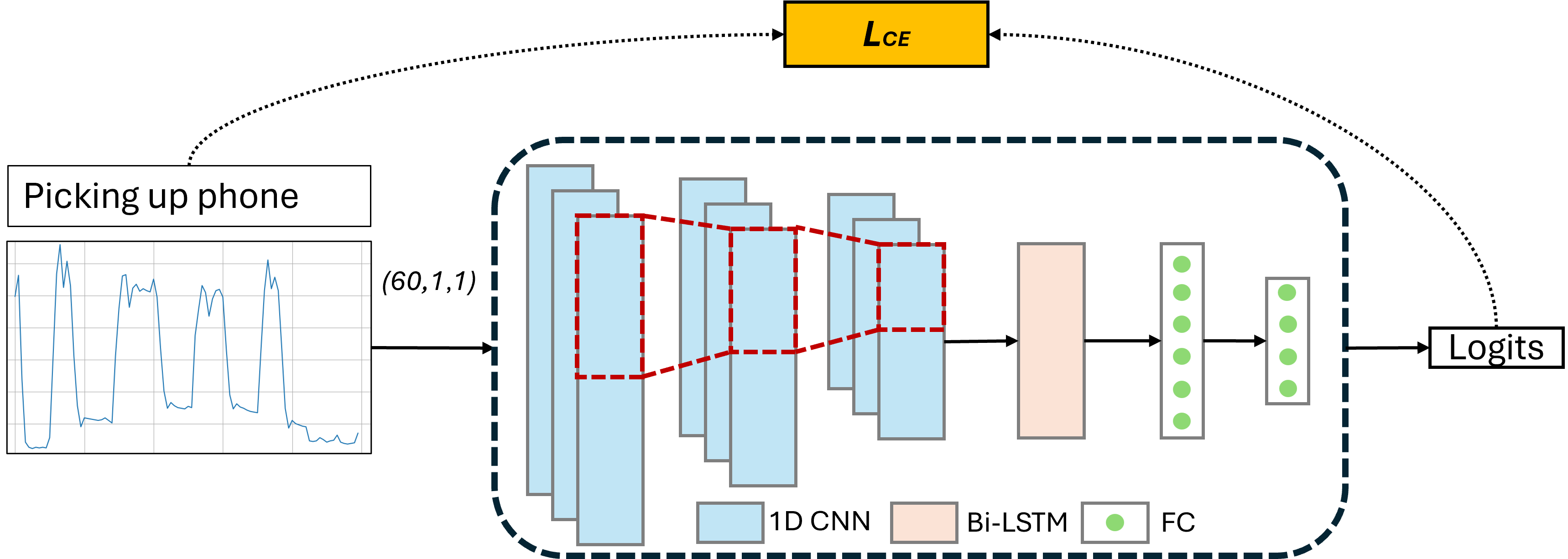}
\caption{Overview of the architecture of LightHAR that uses ALS data only for activity classification.}
\label{fig:a3}
\end{figure}

To train the model, we used the cross-entropy loss function, defined as: 
\begin{equation}
 \mathcal{\text{L}}_{\text{CE}} = -\frac{1}{N}\sum_{i=1}^{N} \sum_{c=1}^{C} y_{i,c} \log(\hat{y}_{i,c})   
\end{equation}
where $N$ is the batch size, $C$ is the number of classes, $y_{i,c}$ is the ground truth probability that sample $i$ belongs to class $c$, and $\hat{y}_{i,c}$ is the predicted probability by the model for class $c$ of sample $i$.

\subsection{Light embedded InertialHAR:}
\label{MLightHAR}
We have designed different strategies for leveraging the knowledge from the ALS modality to enhance the activity recognition accuracy of the IMU modality.
As detailed in the \cref{eva1}, ALS, due to its high sensitivity to external light conditions, is susceptible to environmental noise, especially during significant light changes. In contrast, the accelerometer from IMU is known for its environmental robustness and stability. We've developed a variety of strategies that leverage the unique features of both ALS and IMU sensors. These strategies enable us to build a model that only requires the IMU modality during evaluation, effectively mitigating the impact of environmental noise on ALS. This approach is particularly useful in practical scenarios with substantial light fluctuations.

\subsubsection{MultiLight InertialHAR}

\begin{figure}
\centering
\includegraphics[width=0.9\linewidth]{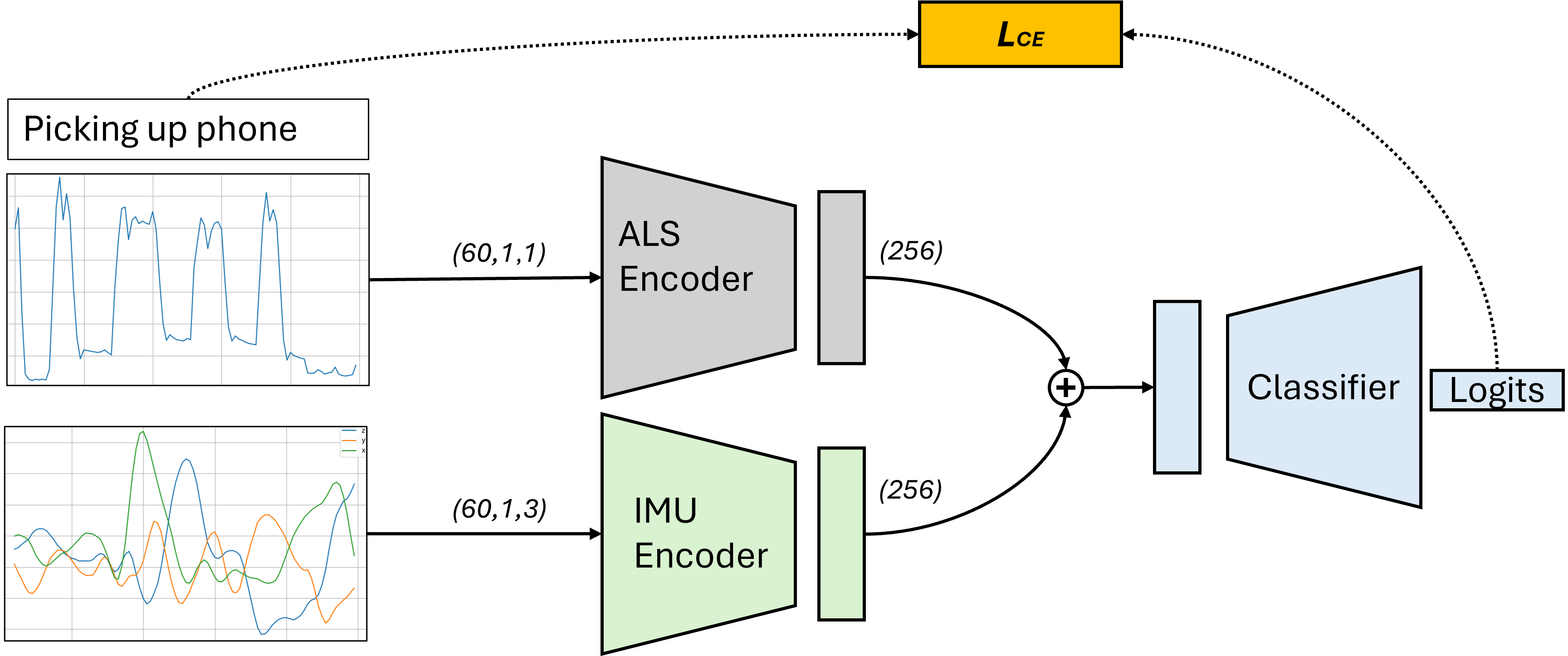}
\caption{Overview of MultiLight InertialHAR that takes both ALS and IMU data during training and relies on IMU only during inference by filling the ALS part with zeros as placeholder during inference.}
\label{fig:a1}
\end{figure}

We designed MultiLight InertialHAR by taking inspiration from classic sensor-fusion models that use more than one modality to improve overall HAR accuracy compared to either unimodal system.
The model contains two encoders: An ALS encoder partially similar to the LightHAR used for activity classification where the full-connected layers are replaced to generate a dense feature vector of size 256.
The IMU encoder also contains a series of 3 1D CNN blocks followed by a bidirectional LSTM and a fully connected layer to generate a dense feature vector of size 256.
The extracted features are concatenated afterward and given to a simple classifier consisting of two fully connected layers to map the intermediate features to an activity class as visualized in Figure \ref{fig:a1}.

Like the LightHAR model, cross-entropy loss was used to train the model. 
MultiLight IneritalHAR processes input ALS data $(N, 1, 1)$, and input accelerometer data $(N, 1, 3)$, where $N$ represents the sequence length, 1 is the feature dimension, and 3 is total channels$(x,y,z)$ to output one of the 10 (9+Null) classes. 

Since we aim to design a HAR system that utilizes both sensor modalities during the training phase and only the IMU modality during the evaluation phase, we develop a unique data pre-processing pipeline to train the model. Each data point in the dataset is converted to 3 instances: the original and instances where one of the two modalities is replaced by zero, enabling us to evaluate the model even when one is unavailable.
During the inference phase, without the presence of ALS data, we can simply $(N, 1, 1)$ input this as a set of 0 and give appropriate data for $(N, 1, 3)$, making it possible to work without changing the architecture.

\subsubsection{ContraLight InertialHAR}
\begin{figure}
\centering
\includegraphics[width=0.9\linewidth]{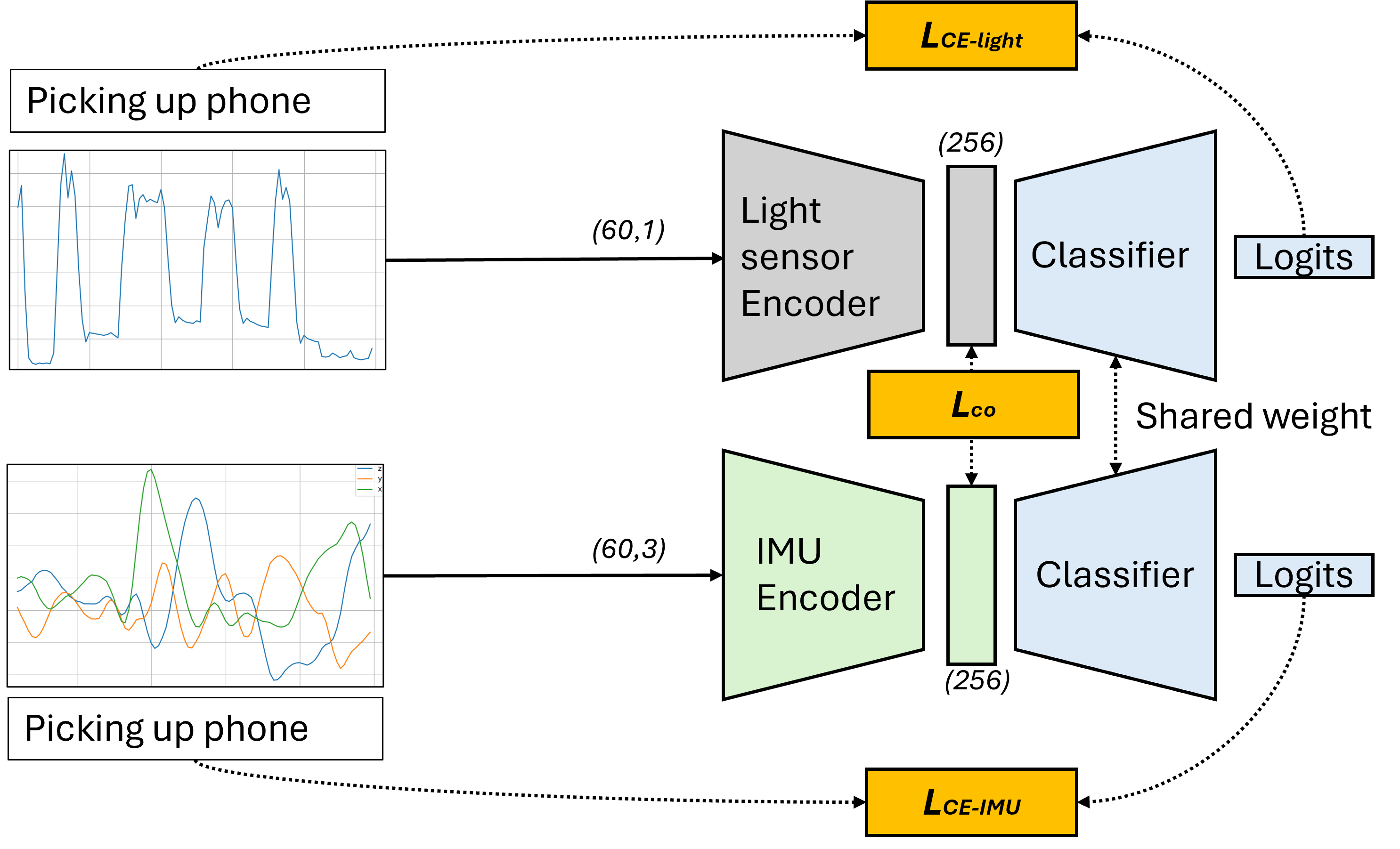}
\caption{Overview of ContraLight InertialHAR that takes both ALS and IMU data during training but only IMU during inference.}
\label{fig:a2}
\end{figure}
Inspired by other multi-modal contrastive learning models, We devised another unique strategy where both light and inertial sensor data are used during the training phase, but only inertial sensor data is used during the evaluation phase by utilizing contrastive learning to train this model. 
The ALS encoder and IMU encoder, identical to those used in MultiLight InertialHAR, extract two feature vectors of size 256. 
Contrastive loss is then applied to both embeddings based on the original classes to bring representations from the same classes closer together.

The contrastive loss $\mathcal{\text{L}}_{\text{co}}$ is defined as:

\begin{equation}
\mathcal{\text{L}}_{\text{co}} = \sum_{i,j} y_{ij} \cdot \max(0, m - \|z_i - z_j\|^2) + (1 - y_{ij}) \cdot \|z_i - z_j\|^2    
\end{equation}
where $z_i$ and $z_j$ are the feature vectors, $y_{ij}$ is a binary label indicating whether $z_i$ and $z_j$ are from the same class, and $m$ is a margin parameter.
Two instances of the same fully connected classifier are utilized with shared weights as visualized in Figure \ref{fig:a2}. 
The overall total loss $\mathcal{L}_{\text{total}}$ is then calculated by summing the contrastive loss and the two cross-entropy losses.

\begin{equation}
    {\text{L}}_{\text{total}} = {\text{L}}_{\text{co}} + {\text{L}}_{\text{CE-light}} + \mathcal{\text{L}}_{\text{CE-IMU}}    
\end{equation}

We used contrastive loss instead of InfoNCE loss as this is a supervised problem. Therefore, we can directly use the labels as individual clusters instead of the self-supervised clustering task, which is useful in cases where the target activities are different from the source activities.  

This approach ensures that during training, the model leverages both $(N, 1, 1)$ ALS and $(N, 1, 3)$ IMU sensor data to learn robust feature representations. However, during inference, we can simply discard the ALS encoder and use the more stable $(N, 1, 3)$ inertial sensor data as input to predict the activity class.
Unlike the MultiLight InertialHAR, we do not need to pass a dummy ALS input, making the model even smaller and more simplified without any significant trade-off.

\section{Evaluation}
\subsection{Training details}
To train the models with the collected data, instances were generated using a sliding window technique with a size of 60 (2 sec) and a step of 15 samples (0.5 sec) for both ALS and IMU sensor data.
The video and extracted pose, which have 24 frames per second (FPS), are first interpolated to make it 30FPS and afterward sliced accordingly to generate a window of size 60 and a step of 15 frames.

All models are trained using a Nvidia A6000 Ada Lovelace GPU and a Ryzen 5900 processor. 
Subjects 1 to 7 constituted the training set, while subjects 8 to 10 formed the test set 1 for ideal light conditions.
Subjects 11 to 13 formed test set 2 for challenging indoor lighting conditions, and subjects 14 to 16 formed test set 3 for outdoor conditions.
Training and validation data were randomly split with a 9:1 ratio during the training process.

The ADAM optimizer, along with a constant learning rate of 0.001, is used to train the model. The models are trained for 300 epochs, and early stopping with a patience of 10 is employed.

\subsection{Unimodal Results}
\label{eva1}
To test the effectiveness of ALS as an activity classification modality, we trained other widely used temporal modalities, such as IMU, Pose, and Video, for activity recognition using the same dataset we collected before.
All modalities interpolated to have the same sampling frequency and step size.
To make them comparable, we made the neural network architecture identical to LightHAR for all other modalities except the input size.
Each model was designed with three CNN blocks for feature extraction, a bi-directional LSTM, and two fully connected layers for activity classification, mirroring the structure of LightHAR.
As stated in \cref{LightHAR}, the input of the LightHar is the ALS signal of size $(60,1,1)$ while the input for InertialHAR is the IMU signal of size $(60,1,3)$. In contrast, the input for the PoseHAR is the extracted SMPL pose from videos using MotionBERT \cite{zhu2023motionbert} of size $(60,22,3)$ with 22 joints, and the input for the VideoHAR is the extracted intermediate video features using Video Vision Transformer (ViViT) \cite{arnab2021vivit} of size $(1,3137,768)$.

\begin{table}
\caption{Classification accuracy, macro F1, total number of learnable parameters, inference time, and number of Floating Point Operation (FLOP) from an ALS, IMU, and vision-based HAR sharing identical architecture for the three different scenarios.}
\label{tab:res1}
\resizebox{\textwidth}{!}{
\begin{tabular}{l|c|c|c|c|c|c}
\hline
Modality & Scenario &Accuracy & Macro F1 & Parameters & Time(ms) & FLOP\\
\hline
ALS & 1 & 0.701$\pm$0.024 & 0.639$\pm$0.039 & & & \\
(LightHAR)& 2 & 0.413$\pm$0.017 & 0.398$\pm$0.021 & \textbf{0.294M} & \textbf{0.292$\pm$0.008} & \textbf{25.050M}\\
& 3 & 0.634$\pm$0.012 & 0.608$\pm$0.029 &  &  & \\
\hline
IMU& 1 & 0.713$\pm$0.041 & 0.657$\pm$0.017 & & & \\
(InertialHAR)& 2 & \textbf{0.706$\pm$0.023} & \textbf{0.688$\pm$0.023} & 0.360M & 0.364$\pm$0.023 & 41.430M\\
& 3 & 0.722$\pm$0.033 & 0.696$\pm$0.017 & & & \\
\hline
Vision& 1 & \textbf{0.913$\pm$0.029} & \textbf{0.896$\pm$0.037} & & & \\
(PoseHAR) & 2 & 0.658$\pm$0.035 & 0.644$\pm$0.017 & 2.425M & 0.503$\pm$0.017 & 557.397M\\
& 3 & \textbf{0.0852$\pm$0.034} & \textbf{0.838$\pm$0.016} & & & \\
\hline
Vision& 1  & 0.517$\pm$0.025 & 0.492$\pm$0.033 & & & \\
(VideoHAR) & 2  & 0.412$\pm$0.024 & 0.396$\pm$0.031 & 10.322M & 0.642$\pm$0.029 & 2531.201M\\
& 3  & 0.366$\pm$0.027 & 0.358$\pm$0.022 & & & \\
\hline
\end{tabular}}
\end{table}

As stated in \cref{tab:res1} we used Accuracy, Macro F1 score, Total number of parameters, Inference time, and Floating Point Operations (FLOP) as metrics to compare all four modalities.
For test sets 1 (indoor with fixed external lights) and 3 (outdoor), PoseHAR performed best, followed by IntertialHAR. LightHAR, despite having the fewest learnable parameters and a single channel input, had comparable results to InertialHAR for test set 1.
For test set 2 (indoor with dynamic external lights), InertialHAR performed best, followed by PoseHAR and LightHAR. This change can be attributed to the accelerometer being light-invariant and more stable than other vision-based modalities.
The VideoHAR, despite having the highest number of learnable parameters, performed worst in all cases, which can be attributed to the feature extracted by ViViT.
ViViT extracts video features but has not been specifically trained to extract the features of the person in the video for activity recognition. The extracted features might contain more information about the background rather than the person himself, which is useless for the activity recognition problem, making it useless for this specific use case.

In terms of inference time, LightHAR, with the lowest number of FLOP, has the fastest inference time, followed by InertialHAR, PoseHAR, and VideoHAR.
PoseHAR and VideoHAR also take intermediate features as input, so considering the inference time for pose estimation and video feature extraction would make this even higher, making them unsuitable for real-time use cases.

Despite LightHAR's promising inference time and comparable results to InertialHAR for test set 1, it does not solve the underlying problem related to the unreliability of ALS sensors in challenging conditions like test sets 1 and 2.
To address this, we developed cross-modal knowledge transfer as described in \cref{MLightHAR}.

\subsection{Cross-modal Results}
\label{eva2}
As discussed in the \cref{eva1}, if we consider all scenarios, InterialHAR is more accurate, more reliable/stable, and has a comparable inference time to LightHAR, making it a better choice for activity recognition.

We developed strategies like MultiLight InertialHAR, which takes $(60,1,3)$ IMU input and a dummy array of $(60,1,1)$ for activity classification. In this strategy, both IMU and light modality from ideal conditions (fixed light indoor) are used for training the model.
Similarly, ContraLight InertialHAR takes only $(60,1,3)$ IMU input for activity classification, but both IMU and light modality from ideal conditions (fixed light indoor) are used for training the model. 
The same metrics from before are used to compare all the models.

\begin{table}
\caption{IMU Classification accuracy, macro F1, total number of learnable parameters, inference time, and number of Floating Point Operations (FLOP) for baseline InertialHAR, Multi-Light InertialHAR, Contra-Light InertialHAR and Sensor Fusion(IMU+ALS).}
\label{tab:res}
\resizebox{\textwidth}{!}{
\begin{tabular}{l|c|c|c|c|c|c}
\hline
Model & Scenario  & Accuracy & Macro F1 & Parameters & Time(ms) & FLOP\\
\hline
InertialHAR& 1 &0.713$\pm$0.041 & 0.657$\pm$0.017 &  &  & \\
(Baseline)& 2 & 0.706$\pm$0.023 & 0.688$\pm$0.023 & \textbf{0.360M} & \textbf{0.363$\pm$0.023} & \textbf{41.430M}\\
& 3 & 0.722$\pm$0.033 & 0.696$\pm$0.017 &  &  & \\
\hline
Multi-Light& 1 & 0.719$\pm$0.035 & 0.681$\pm$0.027 &  &  &  \\
InertialHAR& 2 & 0.711$\pm$0.023 & 0.690$\pm$0.021 & 0.852M & 0.388$\pm$0.022 & 198.216M \\
& 3 & 0.725$\pm$0.034 & 0.696$\pm$0.035 & &  &  \\
\hline
Contra-Light& 1 & 0.755$\pm$0.031 & 0.721$\pm$0.038 &  &  &  \\
InertialHAR& 2 & \textbf{0.731$\pm$0.033} & \textbf{0.719$\pm$0.018} & 0.366M & \textbf{0.363$\pm$0.040} & 41.442M \\
& 3 & \textbf{0.756$\pm$0.018} & \textbf{0.729$\pm$0.029} &  &  &  \\
\hline
\hline
Sensor Fusion& 1 & \textbf{0.858$\pm$0.051} & \textbf{0.820$\pm$0.036} &  &  &  \\
(ALS+IMU)& 2 & 0.681$\pm$0.031 & 0.669$\pm$0.025 & 0.852M & 0.391$\pm$0.031 & 198.216M \\
& 3 &0.723$\pm$0.024 & 0.0693$\pm$0.016 &  &  &  \\
\hline
\end{tabular}}
\end{table}

As we can see in Table \ref{tab:res}, for all 3 test sets, both MultiLight InertialHAR and ContraLight InertialHAR outperformed the baseline InertialHAR although requiring the same input $(60,1,3)$ accelerometer data during inference phase.
The Sensor Fusion model that is identical to MultiLight InertialHAR but requires both $(60,1,3)$ IMU input and $(60,1,1)$ ALS input outperforms all of them in test set 1 having ideal lighting conditions, but its performance gets even worse than the baseline InterialHAR for test set 2 where the lighting conditions are challenging it doesn't provide any improvements for test set 3 either.
Regarding inference time, the baseline InertialHAR, having the lowest number of FLOP, performs faster than all other models.
The MultiLight InertialHAR requires an additional dummy ALS input during inference, which has much higher number of FLOP and is slower than baseline InertialHAR despite being more accurate.
The ContraLight InertialHAR, while not surpassing the baseline, demonstrates a very similar number of FLOP and performs on par with the baseline in terms of speed. This efficiency, combined with its superior accuracy and F1 score compared to both the baseline InertialHAR and MultiLight InertialHAR, makes it a competitive model.

\subsection{Discussions}
As stated in \cref{eva1}, despite having decent inference time and comparable results compared to other sensor modalities like IMUs, ALS can not be used as a universal Unimodal-HAR.
Despite its limited working environments, ALS would make a very good modality for specific use cases in smart indoor environments. For example, hospitals or Care Homes have comparatively stable lighting, and the fast inference, passive sensing, and low-power use of ALS make them suitable for this job.

Also, because of their wide availability in smartphones and smartwatches, they can be used for simple yet repetitive tasks like step counting or other types of fitness activity counters, along with IMU modality through sensor fusion.

As discussed in \cref{eva2}, A large amount of multi-modal activity data can be collected in ideal conditions to enhance other sensor modalities like IMU through our knowledge transfer strategies.

\subsection{Limitations}
In our current study, we exclusively utilized the Galaxy S20 to collect all data, limiting our insights to a single device's performance. Testing a different device type other than the one used to collect the training data could provide valuable insights into cross-device ALS-HAR reliability, particularly in scenarios where other sensor-based modalities like IMU lag behind. Additionally, employing more than one ALS sensor at different parts of the body could potentially enhance overall accuracy. This approach may provide more robustness against varying light conditions and outdoor environments, a direction we plan to explore in future research to improve the robustness and reliability of our findings.

\section{Conclusion}
In summary, our study delves into the realm of wearable ALS for HAR, showcasing its potential in understanding human motions.
We developed LightHAR, a novel approach utilizing wrist-based ambient light signals for HAR, tested it for different scenarios, and compared it with other commonly used modalities for HAR.
By integrating ALS with IMU through sensor fusion and contrastive classification, we enhanced the accuracy of InertialHAR systems. Our light-embedded InertialHAR approach, which relies solely on inertial data during inference, exhibited notable improvements in accuracy compared to traditional IMU-based classifiers.
Although our study is promising, it is essential to acknowledge its limitations, and further research is warranted to validate our findings across devices and explore practical applications of ambient light-enhanced HAR systems.

\subsubsection{Acknowledgements} \label{sec:ack}
The research reported in this paper was supported by the BMBF in the project VidGenSense (01IW21003). 

%\vfill\pagebreak

\bibliographystyle{splncs04}
\bibliography{refs}

\begin{thebibliography}{10}
\providecommand{\url}[1]{\texttt{#1}}
\providecommand{\urlprefix}{URL }
\providecommand{\doi}[1]{https://doi.org/#1}

\bibitem{ahamed2020internet}
Ahamed, F., Shahrestani, S., Cheung, H.: Internet of things and machine learning for healthy ageing: identifying the early signs of dementia. Sensors  \textbf{20}(21), ~6031 (2020)

\bibitem{arnab2021vivit}
Arnab, A., Dehghani, M., Heigold, G., Sun, C., Lu{\v{c}}i{\'c}, M., Schmid, C.: Vivit: A video vision transformer. In: Proceedings of the IEEE/CVF international conference on computer vision. pp. 6836--6846 (2021)

\bibitem{demrozi2020human}
Demrozi, F., Pravadelli, G., Bihorac, A., Rashidi, P.: Human activity recognition using inertial, physiological and environmental sensors: A comprehensive survey. IEEE access  \textbf{8},  210816--210836 (2020)

\bibitem{demrozi2021estimating}
Demrozi, F., Turetta, C., Chiarani, F., Kindt, P.H., Pravadelli, G.: Estimating indoor occupancy through low-cost ble devices. IEEE Sensors Journal  \textbf{21}(15),  17053--17063 (2021)

\bibitem{dontfreeze}
Fortes~Rey, V., Nshimyimana, D., Lukowicz, P.: Don't freeze: Finetune encoders for better self-supervised har. In: Adjunct Proceedings of the 2023 ACM International Joint Conference on Pervasive and Ubiquitous Computing \& the 2023 ACM International Symposium on Wearable Computing. p. 195–196. UbiComp/ISWC '23 Adjunct, Association for Computing Machinery, New York, NY, USA (2023). \doi{10.1145/3594739.3610790}, \url{https://doi.org/10.1145/3594739.3610790}

\bibitem{isuwa2023maximising}
Isuwa, S., Amos, D., Singh, A.K., Al-Hashimi, B.M., Merrett, G.V.: Maximising mobile user experience through self-adaptive content-and ambient-aware display brightness scaling. Journal of Systems Architecture  \textbf{145},  103023 (2023)

\bibitem{jiang2020hand}
Jiang, D., Wu, Y., Demosthenous, A.: Hand gesture recognition using three-dimensional electrical impedance tomography. IEEE Transactions on Circuits and Systems II: Express Briefs  \textbf{67}(9),  1554--1558 (2020)

\bibitem{kumar2023tracking}
Kumar, R.P., Melcher, D., Buttolo, P., Jia, Y.: Tracking occupant activities in autonomous vehicles using capacitive sensing. IEEE Transactions on Intelligent Transportation Systems  (2023)

\bibitem{Lara2013A}
Lara, O.D., Labrador, M.: A survey on human activity recognition using wearable sensors. IEEE Communications Surveys and Tutorials  \textbf{15},  1192--1209 (2013). \doi{10.1109/SURV.2012.110112.00192}

\bibitem{li2019adaptive}
Li, H., Derrode, S., Pieczynski, W.: An adaptive and on-line imu-based locomotion activity classification method using a triplet markov model. Neurocomputing  \textbf{362},  94--105 (2019)

\bibitem{li2018self}
Li, Y., Li, T., Patel, R.A., Yang, X.D., Zhou, X.: Self-powered gesture recognition with ambient light. In: Proceedings of the 31st annual ACM symposium on user interface software and technology. pp. 595--608 (2018)

\bibitem{liu2024imove}
Liu, M., Rey, V.F., Zhang, Y., Ray, L.S.S., Zhou, B., Lukowicz, P.: imove: Exploring bio-impedance sensing for fitness activity recognition. arXiv preprint arXiv:2402.09445  (2024)

\bibitem{martin2020electromyographic}
Mart{\'\i}n-Fuentes, I., Oliva-Lozano, J.M., Muyor, J.M.: Electromyographic activity in deadlift exercise and its variants. a systematic review. PloS one  \textbf{15}(2),  e0229507 (2020)

\bibitem{mcgrath2021upper}
McGrath, J., Neville, J., Stewart, T., Cronin, J.: Upper body activity classification using an inertial measurement unit in court and field-based sports: A systematic review. Proceedings of the institution of mechanical engineers, Part P: Journal of sports engineering and technology  \textbf{235}(2),  83--95 (2021)

\bibitem{mekruksavanich2023human}
Mekruksavanich, S., Jantawong, P., Hnoohom, N., Jitpattanakul, A.: Human activity recognition for people with knee abnormality using surface electromyography and knee angle sensors. In: 2023 Joint International Conference on Digital Arts, Media and Technology with ECTI Northern Section Conference on Electrical, Electronics, Computer and Telecommunications Engineering (ECTI DAMT \& NCON). pp. 483--487. IEEE (2023)

\bibitem{light202}
Mohmed, G., Lotfi, A., Pourabdollah, A.: Employing a deep convolutional neural network for human activity recognition based on binary ambient sensor data. In: Proceedings of the 13th ACM International Conference on PErvasive Technologies Related to Assistive Environments. PETRA '20, Association for Computing Machinery, New York, NY, USA (2020). \doi{10.1145/3389189.3397991}, \url{https://doi.org/10.1145/3389189.3397991}

\bibitem{nguyen2023virtual}
Nguyen, D.A., Pham, C., Le-Khac, N.A.: Virtual fusion with contrastive learning for single sensor-based activity recognition. arXiv preprint arXiv:2312.02185  (2023)

\bibitem{nshimyimana2023contrastive}
Nshimyimana, D., Rey, V.F., Lukowic, P.: Contrastive left-right wearable sensors (imus) consistency matching for har. arXiv preprint arXiv:2311.12674  (2023)

\bibitem{ordonez2016deep}
Ord{\'o}{\~n}ez, F.J., Roggen, D.: Deep convolutional and lstm recurrent neural networks for multimodal wearable activity recognition. Sensors  \textbf{16}(1), ~115 (2016)

\bibitem{perez2021recent}
Perez, A.J., Zeadally, S.: Recent advances in wearable sensing technologies. Sensors  \textbf{21}(20), ~6828 (2021)

\bibitem{pesenti2023imu}
Pesenti, M., Invernizzi, G., Mazzella, J., Bocciolone, M., Pedrocchi, A., Gandolla, M.: Imu-based human activity recognition and payload classification for low-back exoskeletons. Scientific Reports  \textbf{13}(1), ~1184 (2023)

\bibitem{ray2023pressim}
Ray, L.S.S., Zhou, B., Suh, S., Lukowicz, P.: Pressim: An end-to-end framework for dynamic ground pressure profile generation from monocular videos using physics-based 3d simulation. In: 2023 IEEE International Conference on Pervasive Computing and Communications Workshops and other Affiliated Events (PerCom Workshops). pp. 484--489. IEEE (2023)

\bibitem{rey2024enhancing}
Rey, V.F., Ray, L.S.S., Qingxin, X., Wu, K., Lukowicz, P.: Enhancing inertial hand based har through joint representation of language, pose and synthetic imus. arXiv preprint arXiv:2406.01316  (2024)

\bibitem{sadaghiani2023ambient}
Sadaghiani, S.M., Ardakani, A., Bhadra, S.: Ambient light-driven wireless wearable finger patch for monitoring vital signs from ppg signal. IEEE Sensors Journal  (2023)

\bibitem{Salem2022Improved}
Salem, Z., Weiss, A.: Improved spatiotemporal framework for human activity recognition in smart environment. Sensors (Basel, Switzerland)  \textbf{23} (2022). \doi{10.3390/s23010132}

\bibitem{shi2021intelligent}
Shi, C., Li, T., Niu, Q.: An intelligent wallpaper based on ambient light for human activity sensing. In: International Conference on Wireless Algorithms, Systems, and Applications. pp. 441--449. Springer (2021)

\bibitem{vesa2020human}
Vesa, A.V., Vlad, S., Rus, R., Antal, M., Pop, C., Anghel, I., Cioara, T., Salomie, I.: Human activity recognition using smartphone sensors and beacon-based indoor localization for ambient assisted living systems. In: 2020 IEEE 16th International Conference on Intelligent Computer Communication and Processing (ICCP). pp. 205--212. IEEE (2020)

\bibitem{s24113367}
Wang, X., Wang, Y., Wu, J.: Position-aware indoor human activity recognition using multisensors embedded in smartphones. Sensors  \textbf{24}(11) (2024). \doi{10.3390/s24113367}, \url{https://www.mdpi.com/1424-8220/24/11/3367}

\bibitem{xu2022visual}
Xu, C., Li, H., Li, Z., Chen, X., Rathore, A.S., Zhang, H., Wang, K., Xu, W.: The visual accelerometer: A high-fidelity optic-to-inertial transformation framework for wearable health computing. In: 2022 IEEE 10th International Conference on Healthcare Informatics (ICHI). pp. 319--329. IEEE (2022)

\bibitem{yadav2022csitime}
Yadav, S.K., Sai, S., Gundewar, A., Rathore, H., Tiwari, K., Pandey, H.M., Mathur, M.: Csitime: Privacy-preserving human activity recognition using wifi channel state information. Neural Networks  \textbf{146},  11--21 (2022)

\bibitem{zhang2020optosense}
Zhang, D., Park, J.W., Zhang, Y., Zhao, Y., Wang, Y., Li, Y., Bhagwat, T., Chou, W.F., Jia, X., Kippelen, B., et~al.: Optosense: Towards ubiquitous self-powered ambient light sensing surfaces. Proceedings of the ACM on interactive, mobile, wearable and ubiquitous technologies  \textbf{4}(3),  1--27 (2020)

\bibitem{zhang2020data}
Zhang, J., Wu, F., Wei, B., Zhang, Q., Huang, H., Shah, S.W., Cheng, J.: Data augmentation and dense-lstm for human activity recognition using wifi signal. IEEE Internet of Things Journal  \textbf{8}(6),  4628--4641 (2020)

\bibitem{zhou2022quali}
Zhou, B., Suh, S., Rey, V.F., Altamirano, C.A.V., Lukowicz, P.: Quali-mat: Evaluating the quality of execution in body-weight exercises with a pressure sensitive sports mat. Proceedings of the ACM on Interactive, Mobile, Wearable and Ubiquitous Technologies  \textbf{6}(2),  1--45 (2022)

\bibitem{zhu2023motionbert}
Zhu, W., Ma, X., Liu, Z., Liu, L., Wu, W., Wang, Y.: Motionbert: A unified perspective on learning human motion representations. In: Proceedings of the IEEE/CVF International Conference on Computer Vision. pp. 15085--15099 (2023)

\end{thebibliography}

\end{document}